%% file: main.tex
\ifdefined\pdftexversion
\pdfoutput=1
\fi
\documentclass[10pt,twocolumn,letterpaper]{article}

\PassOptionsToPackage{table}{xcolor}
\usepackage[pagenumbers]{cvpr} % To force page numbers, e.g. for an arXiv version
\usepackage{multirow}
\usepackage{booktabs}
\usepackage{makecell}
\usepackage{pifont}

\newcommand{\cmark}{\ding{51}}
\newcommand{\xmark}{\ding{55}}
\input{preamble}
\definecolor{cvprblue}{rgb}{0.21,0.49,0.74}
\usepackage[pagebackref,breaklinks,colorlinks,allcolors=cvprblue]{hyperref}

\def\confName{CVPR}
\def\confYear{2027}

\title{HyperUAV: Local Hypergraph Correspondence for Weakly Aligned RGBT Object Detection in UAV Imagery}

\author{
Ben Liang$^{1}$, Meng Han$^{1}$, Yuan Liu$^{1}$,
Junqi Bai$^{2}$, Hongguang Wei$^{1}$, Xiubao Sui$^{1}$, Qian Chen$^{1}$\\
$^{1}$Nanjing University of Science and Technology\\
$^{2}$The 28th Research Institute of China Electronics Technology Group Corporation
}
\begin{document}
\maketitle
\input{sec/0_abstract}    
\input{sec/1_intro}
% \input{sec/2_formatting}
% \input{sec/3_finalcopy}
{
    \small
    \bibliographystyle{ieeenat_fullname}
    \bibliography{main}
}

\input{sec/X_suppl}

\end{document}

%% file: sec/0_abstract.tex
\begin{abstract}
RGB–Thermal (RGBT) object detection enables reliable UAV perception under challenging conditions, yet spatial misalignment between modalities hinders effective fusion. Such misalignment makes cross-modal correspondence inherently ambiguous. However, existing methods, whether based on offset estimation or cross-modal attention, still match each location in isolation, leaving fusion prone to mismatches and background interference. To address this issue, we propose HyperUAV, a hypergraph-based RGBT detector, which models cross-modal correspondence as sets of plausible local associations rather than isolated matches. Specifically, we introduce a Local Correspondence Hypergraph (LCH), where each hyperedge connects an anchor with its local candidates in the other modality through soft incidences guided by feature affinity, spatial proximity, and shared latent prototypes. Hyperedge messages are delivered to the anchors and propagated back to the candidates, so that overlapping hyperedges allow spatially displaced features to refine one another. Furthermore, we design a Discrepancy-Conditioned Cross-Modulation (DCCM) module that injects complementary information between the re-associated features and adaptively rectifies it with sample- and channel-wise correction strength. Extensive experiments on multiple multimodal UAV datasets demonstrate that our method not only achieves state-of-the-art detection performance but also shows strong robustness to cross-modal misalignment. Code will be publicly released at \url{https://github.com/bloomingvision/HyperUAV}.
\end{abstract}

%% file: sec/1_intro.tex
\section{Introduction}
\label{sec:intro}
RGB--Thermal (RGBT) object detection has attracted increasing attention in unmanned aerial vehicle (UAV) perception due to its ability to exploit complementary information from visible and thermal imagery \citep{sun2022dronevehicle,song2024misaligned,chen2026igianet,yang2026crossweaver}. RGB images provide rich visual details for fine-grained recognition but suffer in poor lighting \citep{cai2023multi,liang2026multi}, whereas illumination-robust thermal images capture thermal radiation yet lack fine texture and sharp boundaries \citep{zhang2025sfnet,shi2025spgfuse,dong2025fusion}. By integrating these complementary characteristics, RGBT detection offers a promising solution for robust object perception across diverse environmental conditions.

\begin{figure}[t]
\centering
\includegraphics[width=\columnwidth]{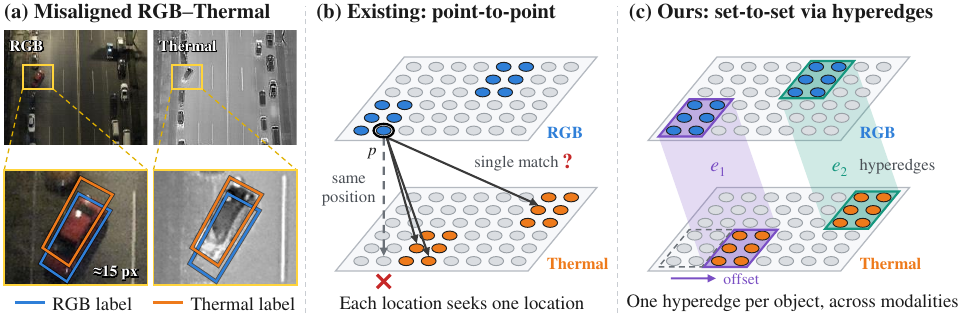}
\caption{Comparison between conventional pairwise fusion and our high-order alignment strategy. Pixel-to-pixel interaction assumes strict coordinate-wise correspondence and is sensitive to weak misalignment. Our method instead models high-order cross-modal correspondences over local support sets, yielding more robust cross-modal interaction.}
\label{fig1:idea}
\end{figure}

In practical systems, paired RGB and thermal images are often only weakly aligned because of differences in sensor viewpoints, fields of view, imaging characteristics, and imperfect spatial calibration \citep{zhang2019weakly,chen2024weakly}. Consequently, features at identical spatial locations may correspond to different object regions or even unrelated background areas. Such discrepancies become particularly problematic for small and densely distributed objects, where slight spatial shifts can lead to substantial feature mismatches. Therefore, simply combining spatially corresponding features may introduce irrelevant cross-modal information and compromise detection performance.

Existing multimodal detectors primarily improve feature fusion by strengthening cross-modal interactions. Attention-based \citep{yuan2024c2former}, Transformer-based \citep{shen2024icafusion}, Mamba-based \citep{dong2025fusion,zhu2025wavemamba}, and frequency-based methods \citep{chen2025mask} enhance feature exchange through spatial-channel reweighting, global context modeling, or modality-specific feature enhancement. Nevertheless, stronger feature interaction does not necessarily guarantee reliable correspondence when the supporting regions of two modalities are spatially displaced. To alleviate this issue, alignment-oriented approaches explore feature alignment, offset-guided sampling, and alignment-free fusion strategies \citep{chen2024weakly,zhang2025wavelet,liu2026cross}. While these methods improve cross-modal feature matching, approaches relying on explicit offset estimation or single-point sampling may remain restrictive when a feature in one modality has multiple plausible supports in the other. Expanding the search region alone is also insufficient, since nearby objects and background structures can produce misleading responses. Therefore, an important question remains: how can cross-modal fusion establish meaningful local correspondences among multiple candidates while preventing unreliable interactions from contaminating feature representations?

To address this challenge, it is necessary to move beyond conventional pairwise interactions and consider relationships involving multiple potential cross-modal correspondences. Hypergraph modeling has been explored in object detection to capture high-order dependencies beyond conventional pairwise interactions. For example, Hyper-YOLO \citep{feng2024hyper} introduces semantic collecting-and-scattering operations to model cross-level and cross-position feature correlations, while YOLOv13 \citep{lei2025yolov13} develops adaptive hyperedge construction for high-order visual reasoning. More recently, hypergraph-based methods have been extended to multimodal detection. M2I2HA \citep{yang2026m2i2ha} employs intra- and inter-modal hypergraph interactions to capture modality-specific and cross-modal relationships, while ProtoHGF-Net \citep{chen2026protohgf} shifts cross-modal interaction from dense spatial features to a compact semantic prototype space, enabling selective prototype-level information exchange. Despite these advances, explicitly characterizing uncertain local correspondences among spatial feature nodes in weakly aligned multimodal inputs remains a distinct and important research problem. In particular, how to represent multiple plausible cross-modal associations while accounting for their spatial and semantic relationships warrants further investigation.

Based on these observations, we propose HyperUAV, a hypergraph-guided framework for weakly aligned RGB--Thermal oriented object detection. HyperUAV consists of two complementary components: Local Correspondence Hypergraph (LCH) and Discrepancy-Conditioned Cross-Modulation (DCCM). First, LCH constructs local hyperedges between RGB and thermal feature nodes. For each anchor, multiple neighboring candidates are associated through soft incidence weights jointly determined by local feature affinity, spatial proximity, and image-conditioned shared-prototype compatibility. The associated candidate features are then aggregated into the anchor frame to form a hyperedge representation encoding the anchor, its cross-modal support, and their discrepancy, while a context-aware hyperedge encoder further incorporates neighboring relation context. Subsequently, DCCM predicts cross-modal discrepancies to guide spatial- and channel-wise feature modulation, enabling selective bidirectional information injection. It further performs discrepancy-conditioned residual correction to refine the modulated features while preserving modality-specific characteristics. Together, LCH establishes local cross-modal correspondence relationships, whereas DCCM regulates cross-modal information exchange and feature refinement. Experiments on DroneVehicle \citep{sun2022dronevehicle}, ESCVehicle \citep{song2026escvehicle}, and ATR-UMOD \citep{chen2025fusion}, demonstrate the effectiveness of HyperUAV, achieving mAP$_{50}$ scores of 83.6\%, 57.1\% and 68.8\%, respectively.

Our main contributions are summarized as follows:
\begin{itemize}
    \item We formulate weakly aligned RGBT fusion as a local cross-modal correspondence reasoning problem and introduce spatially anchored soft hypergraphs to represent multiple plausible cross-modal supports, extending hypergraph modeling toward explicit inter-modal correspondence construction.

    \item We propose Local Correspondence Hypergraph (LCH), which employs image-conditioned shared prototypes to guide local hyperedge incidence estimation. By combining local feature affinity, spatial proximity, and prototype compatibility, LCH constructs adaptive cross-modal relations and performs confidence-regulated bidirectional node--hyperedge--node propagation.

    \item We design Discrepancy-Conditioned Cross-Modulation (DCCM) to adaptively regulate cross-modal feature exchange according to modality discrepancies while retaining modality-specific information.
\end{itemize}
%%%%%%%%%%%%%%%%%%%%%%%%%%%%%%%%%%%%%%%%%%%%%%%%%%%%%%%%%%%%
\section{Related Work}
\subsection{Multimodal Object Detection}
Early object detection methods were designed primarily for RGB imagery \citep{liu2020object,yang2021r3det,han2021align,wei2025multi,liang2026fmc}, but their performance often deteriorates in low-light environments. Since infrared imaging is less sensitive to illumination variations, subsequent studies incorporated infrared data to improve detection accuracy and robustness \citep{zhang2025omni,zhou2025dmm,tang2026adaptive}. To handle weak spatial misalignment, OAFA \citep{chen2024weakly} predicts cross-modal offsets to adaptively align multimodal features before fusion. COMO \citep{liu2026cross} leverages high-level Cross-Mamba interactions to facilitate cross-modal information exchange with improved robustness to spatial misalignment, while offset-guided fusion preserves multiscale representations for effective multimodal detection. CrossWeaver \citep{yang2026crossweaver} introduces a modality-conditioned deformation adapter to adjust spatial sampling fields and combines shared semantics with modality-specific features to reduce alignment-induced quantization errors. Unlike existing methods that address weak spatial alignment through offset estimation and deformable sampling, our method explicitly models cross-modal correspondences using bidirectional, anchor-centered local hypergraphs.

\subsection{Hypergraph Reasoning in Computer Vision}
Graph-based vision models represent image patches as nodes and model their pairwise dependencies through edges. ViG \citep{han2023vision} demonstrates the effectiveness of graph reasoning for visual recognition and detection. Hypergraphs extend conventional graphs by allowing each hyperedge to connect multiple nodes, enabling high-order node--hyperedge--node aggregation \citep{feng2019hypergraph}. This capability has been explored in object detection through Hyper-YOLO \citep{feng2024hyper} and YOLOv13 \citep{lei2025yolov13} for high-order visual feature interaction. HyperDet \citep{li2026hyperdet} adopts an align-then-reason pipeline, applying cross-modal hypergraph reasoning after feature calibration. More recently, M2I2HA \citep{yang2026m2i2ha} and ProtoHGF-Net \citep{chen2026protohgf} have extended hypergraph modeling to multimodal detection by jointly exploring intra-modal dependencies and inter-modal interactions for multimodal feature fusion. Despite these advances, explicitly modeling local cross-modal correspondence uncertainty among spatial feature nodes under weak alignment remains less explored. Different from existing hypergraph-based approaches that focus on high-order context modeling or prototype-level interaction, our method uses hyperedges to explicitly represent local cross-modal correspondence uncertainty before fusion. 
%%%%%%%%%%%%%%%%%%%%%%%%%%%%%%%%%%%%%%%%%%%%%%%%%%%%%%%%%%%%%%%%%%%%%%%%%%%%%%%%%%%%%%%%%%%%%%%%%%%%%%%%%%%%%%%%%%
\section{Method}
\label{sec:method}
\subsection{Overview}
\begin{figure*}[t]
\centering
\includegraphics[width=\textwidth]{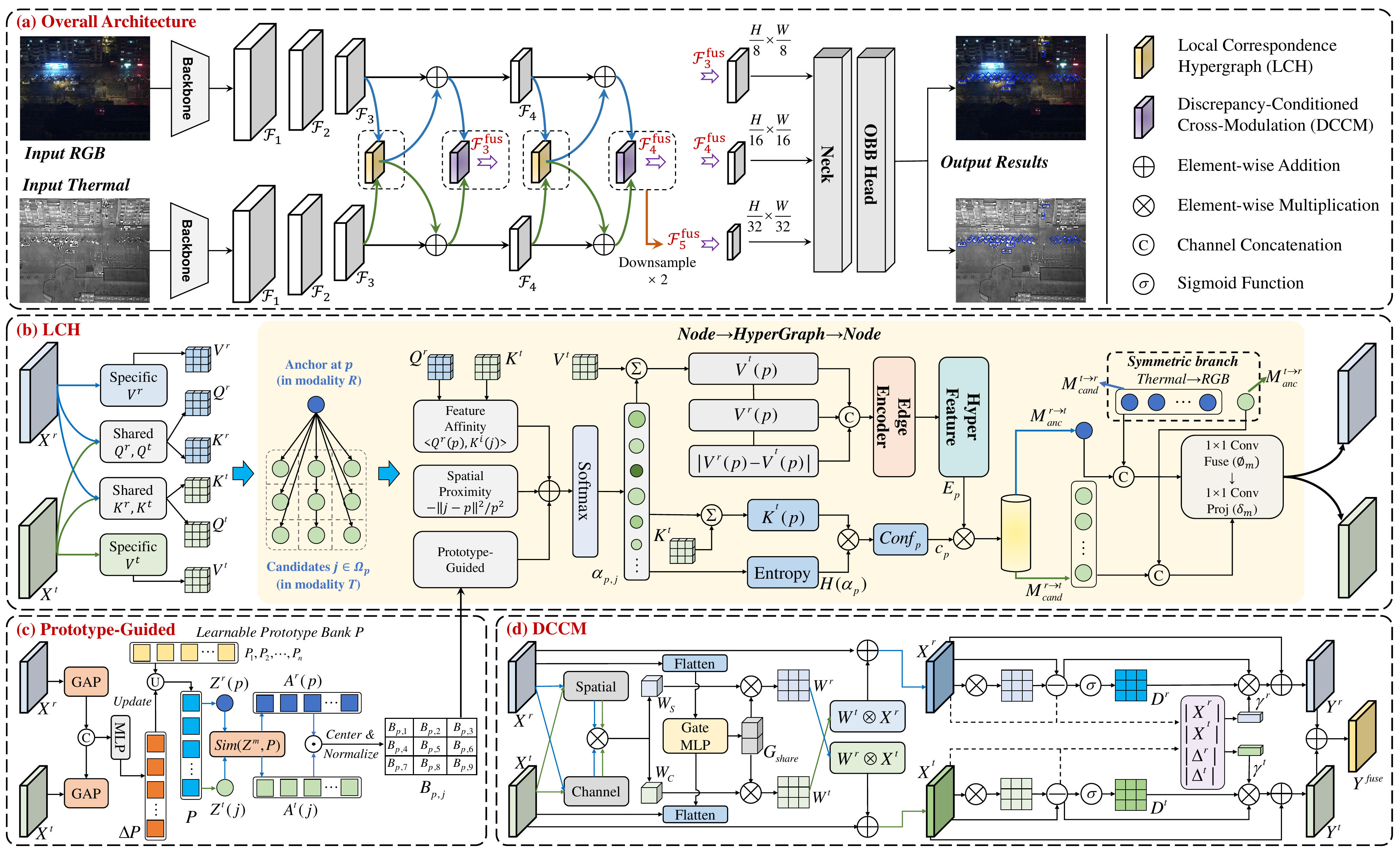}
\caption{Overview of the proposed HyperUAV. At P3 and P4, LCH constructs local cross-modal correspondences and refines the two modality streams before feature addition. DCCM further performs discrepancy-conditioned feature correction.}
\label{fig2:framework}
\end{figure*}

As illustrated in Fig.~\ref{fig2:framework}, we propose HyperUAV for RGBT object detection in UAV imagery. Given paired RGB and thermal images, a dual-branch backbone extracts multi-scale features from each modality. LCH constructs local cross-modal hyperedges by associating each anchor node with multiple neighboring candidates. DCCM further performs discrepancy-conditioned cross-modal modulation and residual correction to exploit complementary information while preserving modality-specific characteristics. The resulting multi-scale representations are fed into the detection head for oriented object detection.

%%%%%%%%%%%%%%%%%%%%%%%%%%%%%%%%%%%%%%%%%%%%%%%%%%%%%%%%%%%%%%%%%%%%%%%%%%%%%%%%%
\subsection{Local Correspondence Hypergraph}
Under weak spatial alignment, corresponding RGB and thermal responses may appear at different feature locations, making direct coordinate-wise fusion unreliable. Instead of explicitly estimating a unique geometric correspondence, we construct local cross-modal hypergraphs that associate each anchor node with multiple neighboring candidates. The resulting node--hyperedge--node interaction consists of two stages: prototype-guided hyperedge construction and confidence-regulated message propagation.

\textbf{Local Hypergraph Initialization.}
Given RGB and thermal features $X^r,X^t\in\mathbb{R}^{C\times H\times W}$ at the same scale, each spatial position is treated as a feature node. We first obtain queries ($Q$) and keys ($K$) through modality-shared projection operators, while using separate value ($V$) projections to retain modality-specific representations. For an anchor node $v_p^a$ at position $p$ in modality $a$, we consider candidate nodes from the other modality $b$ within a $k\times k$ local window $\Omega_k(p)$, where $a,b\in\{r,t\}$ and $a\neq b$. Each anchor defines a local hyperedge whose support consists of the anchor itself and its neighboring cross-modal candidates:
\begin{equation}
e_p^{a\rightarrow b}
=
\{v_p^a\}
\cup
\{v_j^b\mid j\in\Omega_k(p)\}.
\label{eq:pglh_edge_support}
\end{equation}
Accordingly, we define a directed local cross-modal hypergraph
$\mathcal{G}^{a\rightarrow b}
=(\mathcal{V}^a\cup\mathcal{V}^b,\mathcal{E}^{a\rightarrow b})$,
where
$\mathcal{E}^{a\rightarrow b}
=\{e_p^{a\rightarrow b}\}_{p=1}^{HW}$.
Unlike hard correspondence, the candidate nodes are not assumed to contribute equally to an anchor-centered hyperedge. Their participation is determined by adaptive soft incidences, allowing the hypergraph to represent uncertain cross-modal correspondences within a spatially constrained neighborhood.

\textbf{Prototype-Guided Hyperedge Construction.}
Estimating reliable association strengths between an anchor and its candidates is central to local hyperedge construction. While local feature affinity provides fine-grained matching cues and spatial proximity penalizes large displacements, these local signals alone can become ambiguous when RGB and thermal appearances differ drastically. We therefore introduce an image-conditioned shared prototype bank to inject global semantic compatibility into local incidence estimation.

Specifically, both modalities are mapped into a shared semantic space via a common $1\times1$ convolution $f_c$:
\begin{equation}
F_c^m=f_c(X^m)\in\mathbb{R}^{D\times H\times W},
\quad m\in\{r,t\}.
\label{eq:pglh_common_projection}
\end{equation}
We maintain $M$ learnable base prototypes $P_0\in\mathbb{R}^{M\times D}$ ($M=4$, initialized via truncated normal distribution). To dynamically adapt these prototypes to each image pair, the global average pooled ($\operatorname{GAP}$) descriptors of both modalities are concatenated and passed through a two-layer bottleneck $\operatorname{MLP}$ (with a zero-initialized output layer) to predict an instance-specific residual adjustment:
\begin{equation}
\begin{aligned}
g &=
\operatorname{Concat}
\left[
\operatorname{GAP}(F_c^r),
\operatorname{GAP}(F_c^t)
\right]\in\mathbb{R}^{2D},\\
P &=
\operatorname{Norm}
\left(
P_0+\operatorname{Reshape}
(\operatorname{MLP}(g))
\right)\in\mathbb{R}^{M\times D},
\end{aligned}
\label{eq:pglh_prototype_generation}
\end{equation}
where $\operatorname{Norm}(\cdot)$ denotes row-wise $L_2$ normalization so that each prototype $P_h$ has unit length. Each spatial node $p$ is then softly assigned to the $M$ dynamic prototypes according to cosine similarity:
\begin{equation}
A_h^m(p)
=
\frac{
\exp\left(
\tau\langle
\overline F_c^m(p),P_h
\rangle
\right)
}{
\sum_{u=1}^{M}
\exp\left(
\tau\langle
\overline F_c^m(p),P_u
\rangle
\right)
},
\label{eq:pglh_prototype_assignment}
\end{equation}
where $\overline F_c^m(p)=\operatorname{Norm}(F_c^m(p))$ is the unit-normalized feature at position $p$ and $\tau=\operatorname{clip}_{[1,30]}(\exp(\tau_0))$ is a learnable temperature parameter.

For an anchor $p$ and candidate $j\in\Omega_k(p)$, their prototype compatibility is measured by the inner product of their assignment distributions $A^a(p),A^b(j)\in[0,1]^M$:
\begin{equation}
B_{p,j}^{a\rightarrow b}
=
\frac{
\langle A^a(p),A^b(j)\rangle
-1/M
}{
1-1/M
},
\label{eq:pglh_prototype_bias}
\end{equation}
which centers the overlap at the uniform-assignment baseline $1/M$ and scales it by its maximum positive deviation $1-1/M$. We then combine local feature affinity, spatial distance penalty, and prototype compatibility to compute the incidence logit:
\begin{equation}
\ell_{p,j}^{a\rightarrow b}
=
s\langle Q^a(p),K^b(j)\rangle
-
\lambda
\frac{\|j-p\|_2^2}{\max(\rho^2,1)}
+
\eta_c B_{p,j}^{a\rightarrow b},
\label{eq:pglh_incidence_logit}
\end{equation}
where $\rho=(k-1)/2$ is the local search radius, and $\eta_c$ denotes the learnable prototype-guidance strength, initialized at $\eta_c^{(0)}$. Softmax normalization over valid in-bounds candidates in $\Omega_k(p)$ yields the soft incidence weights:
\begin{equation}
\alpha_{p,j}^{a\rightarrow b}
=
\frac{
\exp(\ell_{p,j}^{a\rightarrow b})
}{
\sum_{u\in\Omega_k(p)}
\exp(\ell_{p,u}^{a\rightarrow b})
}.
\label{eq:pglh_soft_incidence}
\end{equation}
These weights specify the relative participation strengths of candidate nodes in the anchor-centered hyperedge $e_p^{a\rightarrow b}$. In terms of the hypergraph preliminaries, the bipartite soft incidence matrix $\widetilde{\mathbf{H}}^{a\rightarrow b}$ is given by
\begin{equation}
\widetilde H_{u,p}^{a\rightarrow b}
=
\begin{cases}
1, & u=v_p^a,\\
\alpha_{p,j}^{a\rightarrow b},
& u=v_j^b,\ j\in\Omega_k(p),\\
0, & \text{otherwise},
\end{cases}
\label{eq:pglh_incidence_matrix}
\end{equation}
Notably, shared prototypes do not broadcast global features across space; instead, they guide local hyperedge construction by re-ranking candidate incidences within $\Omega_k(p)$.

\textbf{Node-to-Hyperedge Aggregation.}
Once the soft incidences have been established, candidate nodes are aggregated into their corresponding anchor-centered hyperedges. For each anchor $p$, we compute the incidence-weighted cross-modal value and key representations:
\begin{equation}
\begin{aligned}
\widehat V^b(p)
&=
\sum_{j\in\Omega_k(p)}
\alpha_{p,j}^{a\rightarrow b}V^b(j),\\
\widehat K^b(p)
&=
\sum_{j\in\Omega_k(p)}
\alpha_{p,j}^{a\rightarrow b}K^b(j).
\end{aligned}
\label{eq:pglh_node_aggregation}
\end{equation}
Here, $\widehat V^b(p)$ collects complementary information from multiple candidate nodes, whereas $\widehat K^b(p)$ retains their aggregated matching representation for subsequent reliability estimation. To encode the relation between the anchor and its cross-modal support, we combine their value representations with their absolute discrepancy:
\begin{equation}
Z^{a\rightarrow b}(p)
=
\operatorname{Concat}
\left[
V^a(p),
\widehat V^b(p),
|V^a(p)-\widehat V^b(p)|
\right].
\label{eq:pglh_edge_input}
\end{equation}
The resulting representations are processed by a hyperedge encoder:
\begin{equation}
E^{a\rightarrow b}
=
\phi_e(Z^{a\rightarrow b}),
\label{eq:pglh_edge_representation}
\end{equation}
where $\phi_e$ consists of a $1\times1$ convolution followed by a $3\times3$ convolution. The first convolution integrates the anchor and its aggregated support, while the second incorporates local context among neighboring anchor-centered representations. Through this node-to-hyperedge aggregation, each hyperedge encodes the joint information of an anchor and its softly associated cross-modal candidates, rather than an isolated pairwise feature interaction.

\textbf{Confidence-Regulated Hyperedge-to-Node Propagation.}
When an anchor lies in background clutter or modality-degraded regions, its local incidence distribution $\alpha_p^{a\rightarrow b}$ may become diffuse or semantically inconsistent. To prevent unreliable hyperedges from contaminating node representations, we construct a diagonal hyperedge weight matrix $\mathbf{W}_e^{a\rightarrow b}=\operatorname{diag}(c_1^{a\rightarrow b},\ldots,c_{HW}^{a\rightarrow b})$ based on incidence concentration and semantic agreement.

First, incidence concentration is quantified via normalized Shannon entropy:
\begin{equation}
c_{\mathrm{ent}}(p)
=
\operatorname{clip}_{[0,1]}
\left(
1-
\frac{
\mathcal H(\alpha_p^{a\rightarrow b})
}{
\max(\log(k^2),\epsilon)
}
\right),
\label{eq:pglh_entropy_conf}
\end{equation}
where
\begin{equation}
\mathcal H(\alpha_p)
=
-\sum_{j\in\Omega_k(p)}
\bar\alpha_{p,j}
\log\bar\alpha_{p,j},
\quad
\bar\alpha_{p,j}
=
\max(\alpha_{p,j},\epsilon).
\label{eq:pglh_entropy}
\end{equation}
A sharply peaked incidence distribution yields $c_{\mathrm{ent}}(p)\rightarrow 1$, whereas a uniform, uncertain distribution yields $c_{\mathrm{ent}}(p)\rightarrow 0$. Second, we measure post-aggregation semantic agreement between the unit-norm anchor query $Q^a(p)$ and the $L_2$-normalized aggregated key $\operatorname{Norm}(\widehat K^b(p))$:
\begin{equation}
c_{\mathrm{agr}}(p)
=
\operatorname{clip}_{[0,1]}
\left(
\frac{
1+
\langle
Q^a(p),
\operatorname{Norm}(\widehat K^b(p))
\rangle
}{2}
\right).
\label{eq:pglh_agreement}
\end{equation}
The composite hyperedge confidence is defined as
\begin{equation}
c_p^{a\rightarrow b}
=
c_{\mathrm{ent}}(p)
c_{\mathrm{agr}}(p)\in[0,1],
\label{eq:pglh_confidence}
\end{equation}
which modulates the hyperedge representation prior to node redistribution:
\begin{equation}
\widetilde E_p^{a\rightarrow b}
=
c_p^{a\rightarrow b}E_p^{a\rightarrow b}.
\label{eq:pglh_gated_edge}
\end{equation}
Hyperedge-to-node propagation is then performed along two paths. The anchor node directly receives its corresponding hyperedge message:
\begin{equation}
M_{\mathrm{anc}}^{a\rightarrow b}(p)
=
\widetilde E_p^{a\rightarrow b}.
\label{eq:pglh_anchor_message}
\end{equation}
Meanwhile, each candidate node receives the accumulated messages from all incident anchor-centered hyperedges:
\begin{equation}
M_{\mathrm{cand}}^{a\rightarrow b}(j)
=
\frac{
\sum\nolimits_{p:\,j\in\Omega_k(p)}
\alpha_{p,j}^{a\rightarrow b}\widetilde E_p^{a\rightarrow b}
}{
\max\left(
\sum\nolimits_{p:\,j\in\Omega_k(p)}
\alpha_{p,j}^{a\rightarrow b},\epsilon
\right)
}.
\label{eq:pglh_candidate_message}
\end{equation}
The incidence-based normalization prevents the accumulated message magnitude from growing excessively when a candidate participates in multiple overlapping hyperedges. Consequently, information aggregated from a local cross-modal support set can contribute to multiple spatial nodes through their shared hyperedge associations.

We independently perform the above node--hyperedge--node propagation in both RGB-to-thermal and thermal-to-RGB directions. For each modality, its anchor-side messages are combined with the candidate-side messages received from the opposite directional hypergraph:
\begin{equation}
Y^a
=
X^a+
\Delta_a
\left(
f_{\mathrm{fuse}}^a
\left(
\operatorname{Concat}
\left[
M_{\mathrm{anc}}^{a\rightarrow b},
M_{\mathrm{cand}}^{b\rightarrow a}
\right]
\right)
\right),
\label{eq:pglh_feature_update}
\end{equation}
where $f_{\mathrm{fuse}}^a$ denotes a modality-specific $1\times1$ fusion convolution and $\Delta_a$ is the output projection. In this way, LCH integrates prototype-conditioned local relation construction with confidence-regulated bidirectional propagation, allowing cross-modal features to interact without requiring a unique coordinate-level correspondence.
%%%%%%%%%%%%%%%%%%%%%%%%%%%%%%%%%%%%%%%%%%%%%%%%%%%%%%%%%%%%%%%%%%%%%%
\subsection{Discrepancy-Conditioned Cross-Modulation}
LCH tolerates local mismatch, but the two streams still differ in their response statistics: the discrepancy that remains after interaction varies across samples, channels, and positions, so a fixed correction cannot balance modality-specific preservation against cross-modal calibration. DCCM therefore derives the modulation factors from the post-interaction discrepancy itself, so that the correction is conditioned on how strongly, and where, the two modalities disagree.

\textbf{Reliability Mask Estimation.}
Given $Y^r$ and $Y^t$, we estimate modality-specific spatial weights $W_s^m\in(0,1)^{1\times H\times W}$, channel weights $W_c^m\in(0,1)^{C\times 1\times 1}$, and a shared pixel-wise gate $G_{\mathrm{pix}}\in(0,1)^{C\times H\times W}$ ($m\in\{r,t\}$) from joint bimodal cues. Specifically, channel-wise mean and max maps of both modalities are concatenated into a 4-channel descriptor and passed through a two-layer $1\times 1$ convolutional network with sigmoid activation to produce $W_s^r,W_s^t$. Concurrently, global average and max pooled descriptors of $\operatorname{Concat}[Y^r,Y^t]\in\mathbb{R}^{2C\times H\times W}$ are fed into a shared bottleneck MLP and summed before a sigmoid gate to obtain $W_c^r,W_c^t$. Combined with the pixel-wise interaction gate $G_{\mathrm{pix}}=\sigma(\operatorname{MLP}_g(\operatorname{Concat}[Y^r,Y^t]))$, the composite reliability masks are synthesized as:
\begin{equation}
I^m = G_{\mathrm{pix}}\odot W_c^m \odot W_s^m, \quad m\in\{r,t\},
\label{eq:dccm_mask}
\end{equation}
where $\odot$ denotes element-wise multiplication with broadcasting.

\textbf{Cross-Modal Injection.}
Each modality first receives gated complementary evidence from the other:
\begin{equation}
\widetilde{Y}^r = Y^r + I^r\odot Y^t,
\qquad
\widetilde{Y}^t = Y^t + I^t\odot Y^r.
\label{eq:dccm_injection}
\end{equation}
The mask is indexed by the receiving stream: $I^r$ weights the thermal evidence admitted into the RGB stream, and vice versa. A direct way to suppress background clutter in $\widetilde{Y}^m$ is to output the strictly gated target $\widehat{Y}^m = I^m\odot\widetilde{Y}^m$. This, however, over-attenuates features when an object is salient in only one modality, because $I^m$ also suppresses positions with weak cross-modal support. We therefore treat strict gating as a residual correction,
$\rho^m = \widehat{Y}^m-\widetilde{Y}^m = (I^m-\mathbf{1})\odot\widetilde{Y}^m$,
and learn how much of it to apply.

\textbf{Discrepancy-Conditioned Calibration.}
The correction strength is predicted jointly for the two streams from their magnitudes and residuals:
\begin{equation}
[\gamma^r,\gamma^t]
=
2\sigma\!\Big(
\operatorname{MLP}_\gamma\!\big(
\operatorname{GAP}\big(\,[\,|\widetilde{Y}^r|\,\|\,|\widetilde{Y}^t|\,\|\,|\rho^r|\,\|\,|\rho^t|\,]\,\big)
\big)
\Big),
\label{eq:dccm_gamma}
\end{equation}
with $\gamma^m\in(0,2)^{C}$. The last layer of $\operatorname{MLP}_\gamma$ is zero-initialized, so $\gamma^m=\mathbf{1}$ at the beginning of training. A local magnitude gate $D^m=\sigma(|\rho^m|)$ further makes the correction follow the residual strength, and the calibrated feature is
\begin{equation}
\begin{aligned}
\bar{Y}^m
&= \widetilde{Y}^m + \gamma^m \odot D^m \odot \rho^m,
\end{aligned}
\label{eq:dccm_final}
\end{equation}
where $\omega^m=\gamma^m\odot D^m$. The calibrated features are summed element-wise, $Y^{\mathrm{fuse}} = \bar{Y}^r + \bar{Y}^t$, and passed to the downstream detection layers.
%%%%%%%%%%%%%%%%%%%%%%%%%%%%%%%%%%%%%%%%%%%%%%%%%%%%%%%%%%%%%%%
\section{Experiments}
\begin{table*}[!t]
\centering
\caption{Quantitative comparison with state-of-the-art methods on the
DroneVehicle test set. All methods are trained and evaluated at an input resolution of $640\times640$. $\dagger$ denotes results reproduced under the
same training and evaluation settings.}
\label{tab1:dronevehicle}
\small
\setlength{\tabcolsep}{1.5pt}
\renewcommand{\arraystretch}{0.8}
\begin{tabular}{lccccccccccc}
\toprule
Method & Pub. & Modality
& Car & Truck & Bus & Van & Freight car
& mAP$_{50}$ & mAP
& Params (M) & FLOPs (G) \\
\midrule
R$^3$Det \citep{yang2021r3det} & AAAI'21
& \multirow{4}{*}{RGB}
& 80.3 & 56.1 & 80.2 & 44.4 & 42.7
& 60.8 & 27.4
& 73.8 & 173.0 \\
S$^2$ANet \citep{han2021align} & TGRS'21
& & 80.0 & 54.2 & 84.9 & 43.8 & 42.2
& 61.0 & 31.4
& 70.7 & 120.6 \\
KFIoU \citep{yang2022kfiou} & ICLR'23
& & 76.3 & 50.0 & 78.2 & 35.7 & 33.2
& 54.7 & 28.0
& 41.9 & 131.2 \\
PKINet-S \citep{cai2024poly} & CVPR'24
& & 76.4 & 53.0 & 79.1 & 45.7 & 40.8
& 59.0 & 31.2
& -- & -- \\
\midrule
R$^3$Det \citep{yang2021r3det} & AAAI'21
& \multirow{4}{*}{T}
& 89.5 & 48.3 & 87.1 & 39.9 & 16.6
& 62.3 & 36.7
& 73.8 & 173.0 \\
S$^2$ANet \citep{han2021align} & TGRS'21
& & 89.9 & 54.5 & 88.9 & 48.4 & 55.8
& 67.5 & 40.4
& 70.7 & 120.6 \\
KFIoU \citep{yang2022kfiou} & ICLR'23
& & 90.2 & 65.0 & 89.0 & 43.0 & 48.5
& 67.1 & 41.2
& 41.9 & 131.2 \\
PKINet-S \citep{cai2024poly} & CVPR'24
& & 90.2 & 67.3 & 88.8 & 51.2 & 56.3
& 70.8 & 43.8
& -- & -- \\
\midrule
CALNet \citep{he2023multispectral} & ACMMM'23
& & 90.3 & 76.2 & 89.1 & 58.5 & 63.0
& 75.4 & 48.2
& 39.5 & 52.9 \\
C$^2$Former \citep{yuan2024c2former} & TGRS'24
& \multirow{13}{*}{RGB+T}
& 90.2 & 68.3 & 89.8 & 58.5 & 64.4
& 74.2 & 47.5
& 118.5 & -- \\
OAFA \citep{chen2024weakly} & CVPR'24
& & 90.1 & 75.4 & 89.8 & 61.8 & 68.2
& 77.1 & 50.1
& -- & -- \\
DMM-S$^2$ANet \citep{zhou2025dmm} & TGRS'25
& & 90.4 & 79.8 & 89.9 & 68.6 & 68.2
& 79.4 & 52.1
& 88.0 & -- \\
GAANet \citep{zheng2025gaanet} & TII'25
& & 90.3 & 80.5 & 89.8 & \textbf{68.7} & 65.6
& 79.0 & 52.1
& 79.0 & 194.9 \\
Fusion-Mamba \citep{dong2025fusion} & TMM'25
& & 96.7 & 80.2 & 95.3 & 64.0 & 59.8
& 79.2 & 56.0
& 287.6 & -- \\
WaveMamba \citep{zhu2025wavemamba} & ICCV'25
& & 95.0 & 80.4 & 90.6 & 64.5 & 68.5
& 79.8 & 60.5
& 69.1 & -- \\
M$^2$D-LIF \citep{zhao2025rethinking} & ICCV'25
& & 97.8 & 81.0 & 96.0 & 64.6 & 67.9
& 81.4 & \underline{68.1}
& 37.1 & -- \\
HyperDet $^\dagger$ \citep{li2026hyperdet} & JKSU-CIS'26
& & \underline{98.5} & 80.8 & 96.0 & 62.4 & 68.3
& 81.2 & 65.6
& \textbf{3.2} & \textbf{9.4} \\
COMO $^\dagger$ \citep{liu2026cross} & IF'26
& & 97.8 & 81.3 & 94.8 & 63.9 & 68.4
& 81.2 & 66.7
& 24.6 & 90.7 \\
CrossWeaver-N $^\dagger$ \citep{yang2026crossweaver} & CVPR'26
& & 98.4 & 79.2 & 95.0 & 63.2 & 67.3
& 80.6 & 57.3
& \underline{5.4} & 19.5 \\
ProtoHGF-Net $^\dagger$ \citep{chen2026protohgf} & ACMMM'26
& & \underline{98.5} & 82.4 & 95.7 & 63.8 & 68.0
& 81.7 & 66.8
& 51.9 & 136.5 \\
\rowcolor{blue!10}
HyperUAV-N (Ours) & --
& & \underline{98.5} & \underline{82.8} & \underline{96.4} & 66.6 & \underline{70.0}
& \underline{82.9} & 67.4
& \textbf{3.2} & \underline{9.6} \\
\rowcolor{blue!10}
HyperUAV-S (Ours) & --
& & \textbf{98.6} & \textbf{83.9} & \textbf{96.6} & 67.4 & \textbf{71.6}
& \textbf{83.6} & \textbf{69.0}
& 11.3 & 31.7 \\
\bottomrule
\end{tabular}
\end{table*}

\subsection{Datasets}
We evaluate our method on three UAV-based RGB--IR object detection benchmarks: 
\textbf{(1) DroneVehicle} \citep{sun2022dronevehicle} contains 28,439 image pairs (17,990 train, 1,469 validation, and 8,980 test) annotated with oriented bounding boxes across five categories (\textit{car}, \textit{truck}, \textit{bus}, \textit{van}, and \textit{freight car}). 
\textbf{(2) ESCVehicle} \citep{song2026escvehicle} comprises 6,983 training, 507 validation, and 3,237 testing pairs across seven vehicle categories and nine scenes spanning daytime, nighttime, and adverse weather. 
\textbf{(3) ATR-UMOD} \citep{chen2025fusion} provides 13,353 image pairs (11,850 train and 1,503 test) across 11 categories, captured at 80--300\,m altitudes and $0^\circ$--$75^\circ$ angles with six annotated condition attributes (altitude, angle, time, weather, illumination, and scenario).

\subsection{Experimental Settings}
All experiments are conducted on an NVIDIA RTX 4090D GPU. Following the general architectural design of the representative one-stage detector YOLO11, we develop a dual-stream multimodal detection architecture. All input images are resized to $640\times640$, and models are optimized using SGD with a learning rate of 0.01, a momentum of 0.937, and a weight decay of $5\times10^{-4}$. Automatic mixed precision is disabled, and deterministic training is adopted to ensure fair and reproducible comparisons. All models are trained from scratch without pretrained weights, using a batch size of 16 for 150, 100, and 150 epochs on DroneVehicle, ATR-UMOD, and ESCVehicle, respectively. We follow the official evaluation protocols of each dataset, using the corresponding mean Average Precision (mAP) metrics for performance evaluation.
%%%%%%%%%%%%%%%%%%%%%%%%%%%%%%%%%%%%%%%%%%%%%
\subsection{Main Results}
\noindent\textbf{Comparison Results on DroneVehicle.}
We provide a comprehensive comparison between our HyperUAV and existing state-of-the-art methods on the DroneVehicle dataset, with results detailed in Table~\ref{tab1:dronevehicle}. HyperUAV-S achieves the best overall performance, reaching 83.6\% mAP$_{50}$ and 69.0\% mAP, while maintaining only 11.3M parameters and 31.7G FLOPs. It also obtains strong class-wise results, particularly for Truck, Bus, Van, and Freight car. Furthermore, HyperUAV-N provides a better balance between detection accuracy and computational efficiency.

\noindent\textbf{Comparison Results on ESCVehicle.}
Table~\ref{tab:escvehicle_overall_comparison} reports the results on ESCVehicle. Notably, HyperUAV-N achieves 54.4\% mAP$_{50}$ and 40.2\% mAP, surpassing C$^2$-VeD \citep{song2026escvehicle} by 2.0 and 1.6 percentage points, respectively, with only 3.2M parameters and 9.6G FLOPs. HyperUAV-S further improves mAP to 43.9\%, achieving the best overall accuracy.

\noindent\textbf{Comparison Results on ATR-UMOD.}
Table~\ref{tab:atrumod_overall_comparison} reports the comparison on ATR-UMOD. Overall, our method achieves the highest mAP of 68.8\%, outperforming the previous best PCDF by 5.5 percentage points. It also achieves the best results across all five weather conditions, with particularly notable improvements under foggy conditions (W2), demonstrating its effectiveness under diverse weather scenarios.

\begin{table}[!t]
\centering
\caption{Quantitative comparison with state-of-the-art methods on the
ESCVehicle test set.}
\label{tab:escvehicle_overall_comparison}
\small
\setlength{\tabcolsep}{0.8pt}
\renewcommand{\arraystretch}{0.8}
\begin{tabular}{lccccc}
\toprule
Method & Modality & mAP$_{50}$ & mAP
& Params (M)  & FLOPs (G)  \\
\midrule
RoI Transformer \citep{ding2019learning}
& RGB & 37.2 & 19.3 & 87.2 & 148.7 \\
S$^2$ANet \citep{han2021align}
& RGB & 35.5 & 17.5 & 70.7 & 120.6 \\
Oriented R-CNN \citep{xie2021oriented}
& RGB & 39.3 & 20.6 & 73.2 & 134.8 \\
KLD \citep{yang2021learning}
& RGB & 32.9 & 16.2 & 41.9 & 131.2 \\
\midrule
RoI Transformer \citep{ding2019learning}
& T & 22.0 & 10.5 & 87.2 & 148.7 \\
S$^2$ANet \citep{han2021align}
& T & 21.1 & 9.7 & 70.7 & 120.6 \\
Oriented R-CNN \citep{xie2021oriented}
& T & 24.3 & 11.4 & 73.2 & 134.8 \\
OSIV-Net \cite{zhang2025omni}
& T & 37.0 & 19.1 & \underline{10.8} & 29.38 \\
\midrule
ICAFusion \citep{shen2024icafusion}
& RGB+T & 45.0 & 30.3 & 24.5 & 39.2 \\
CALNet \citep{he2023multispectral}
& RGB+T & 48.8 & 31.8 & 39.5 & 52.9 \\
C$^2$Former \citep{yuan2024c2former}
& RGB+T & 48.7 & 30.8 & 36.8 & 80.3 \\
CMA-Det \citep{song2024misaligned}
& RGB+T & 47.6 & 27.9 & 28.9 & 39.8 \\
C$^2$-VeD \citep{song2026escvehicle}
& RGB+T & 52.4 & 38.6 & 16.5 & \underline{13.1} \\
\rowcolor{blue!10}
HyperUAV-N
& RGB+T & \underline{54.4} & \underline{40.2}
& \textbf{3.2} & \textbf{9.6} \\
\rowcolor{blue!10}
HyperUAV-S
& RGB+T & \textbf{57.1} & \textbf{43.9}
& 11.3 & 31.7 \\
\bottomrule
\end{tabular}
\end{table}
%%%%%%%%%%%%%%%%%%%%%%%%%%%%%%%%%%%%

%%%%%%%%%%%%%%%%%%%%%%%%%%%%%%%%%%%%%%%%%%%%%%%%%%%%%%%%%%%%%%%
\subsection{Ablation Analysis}
\noindent\textbf{Effectiveness of the Proposed Module.}
Table~\ref{tab:ablation_components} reports the component ablations on DroneVehicle and ATR-UMOD. CrossP5 retains dual-stream feature extraction at P3/P4 while generating a shared single-stream P5 by downsampling the fused P4 features, reducing redundant high-level computation. Compared with the original baseline, this configuration reduces Params from 3.9M to 2.6M and FLOPs from 9.6G to 7.8G, while maintaining or improving detection accuracy on both datasets. Individually, LCH improves mAP by 1.1 and 1.3 percentage points on DroneVehicle and ATR-UMOD, respectively, while DCCM brings gains of 0.8 and 0.5 points. Combining both modules achieves the best performance, reaching 67.4\% and 48.4\% mAP on the two datasets, respectively. Notably, the complete model requires only 3.2M parameters and 9.6G FLOPs, demonstrating the effectiveness of the proposed modules with limited computational overhead.

\begin{table}[!t]
\centering
\caption{Quantitative comparison with state-of-the-art methods on the ATR-UMOD test set. W1--W5 denote Cloudy, Foggy, Snowy, Sunny, and Night,
respectively.}
\label{tab:atrumod_overall_comparison}
\small
\setlength{\tabcolsep}{6pt}
\renewcommand{\arraystretch}{0.8}
\begin{tabular}{lcccccc}
\toprule
Method & mAP & W1 & W2 & W3 & W4 & W5 \\
\midrule
CALNet \citep{he2023multispectral}
& 53.8 & 54.8 & 28.8 & 38.2 & 48.7 & 32.5 \\
OAFA \citep{chen2024weakly}
& 57.9 & 62.1 & 30.9 & 48.4 & 54.6 & 38.0 \\
YOLOrs \citep{sharma2020yolors}
& 60.0 & 69.6 & 33.2 & 50.1 & 57.1 & 35.4 \\
PCDF \citep{chen2025fusion}
& 63.3 & 71.2 & 32.6 & 53.4 & 60.1 & 41.3 \\
\rowcolor{blue!10}
HyperUAV-S
& \textbf{68.8}
& \textbf{78.1}
& \textbf{38.8}
& \textbf{56.0}
& \textbf{64.7}
& \textbf{44.8} \\
\bottomrule
\end{tabular}
\end{table}
\noindent\textbf{Deployment Strategy.}
Table~\ref{tab:deploy_ablation_deployment} compares different deployment configurations of LCH and DCCM. With both modules enabled, restricting their deployment to P3/P4 and generating P5 by downsampling the fused P4 features (CrossP5) improves mAP$_{50}$/mAP from 65.4\%/48.0\% to 66.4\%/48.4\% compared with the P3/P4/P5 configuration. Meanwhile, Params and FLOPs decrease from 4.7M/11.1G to 3.2M/9.6G. This suggests that P3/P4 provide sufficient spatial details for cross-modal interaction, while the fused P4 features already provide integrated semantic information for subsequent P5 representation learning, making additional P5-level fusion unnecessary.
\begin{figure}[!t]
\centering
\includegraphics[width=\columnwidth]{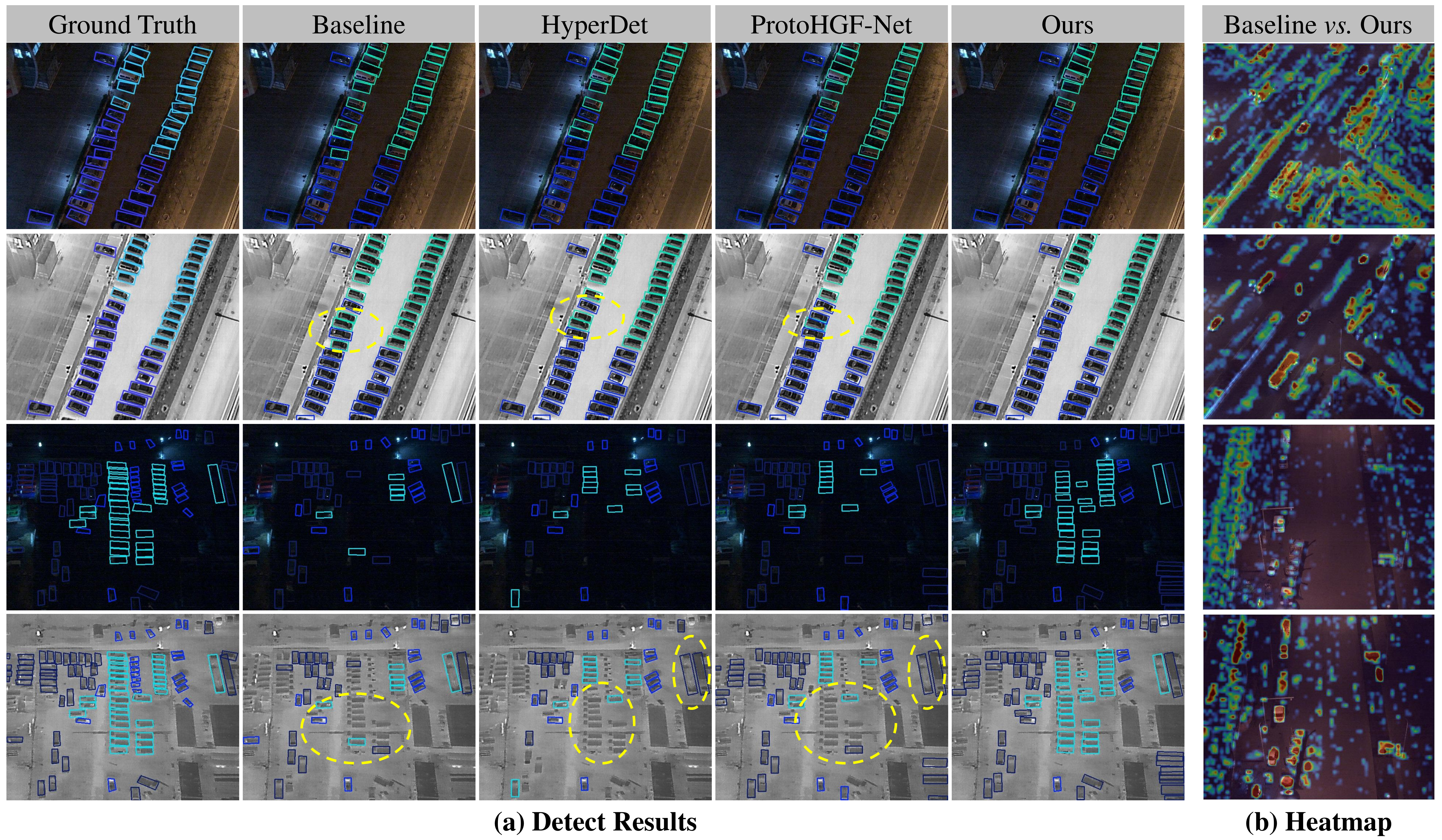}
\caption{Qualitative comparison of the proposed method against other competing methods under complex aerial scenarios.}
\label{fig:detect}
\end{figure}
% ----------------------- Table1 -------------------- %
\begin{table}[t]
\centering
\caption{\textbf{Component ablations on DroneVehicle and ATR-UMOD.}
Baseline denotes the original dual-stream model.}
\label{tab:ablation_components}
\small
\setlength{\tabcolsep}{2pt}
\renewcommand{\arraystretch}{0.8}
\begin{tabular*}{\columnwidth}{@{\extracolsep{\fill}}lcccccc@{}}
\toprule
Deployment & LCH & DCCM & mAP$_{50}$ & mAP & Params & FLOPs \\ 
\midrule

\multicolumn{7}{@{}l}{\textit{\textbf{DroneVehicle Dataset}}} \\
\addlinespace[1pt]
Baseline
& \xmark & \xmark & 80.9 & 65.4 & 3.9 & 9.6 \\
\cmidrule{1-7}

\multirow{4}{*}{\makecell[l]{P3/P4\\(CrossP5)}}
& \xmark & \xmark & 81.1 & 65.5 & 2.6 & 7.8 \\
& \cmark & \xmark & 82.0 & 66.6 & 3.1 & 9.3 \\
& \xmark & \cmark & 81.9 & 66.3 & 3.0 & 8.3 \\
& \cellcolor{blue!10}\cmark
& \cellcolor{blue!10}\cmark
& \cellcolor{blue!10}\textbf{82.9}
& \cellcolor{blue!10}\textbf{67.4}
& \cellcolor{blue!10}3.2
& \cellcolor{blue!10}9.6 \\
\midrule

\multicolumn{7}{@{}l}{\textit{\textbf{ATR-UMOD Dataset}}} \\
\addlinespace[1pt]
Baseline
& \xmark & \xmark & 62.2 & 45.3 & 3.9 & 9.6 \\
\cmidrule{1-7}

\multirow{4}{*}{\makecell[l]{P3/P4\\(CrossP5)}}
& \xmark & \xmark & 64.2 & 46.6 & 2.6 & 7.8 \\
& \cmark & \xmark & 65.4 & 47.9 & 3.1 & 9.3 \\
& \xmark & \cmark & 64.7 & 47.1 & 3.0 & 8.3 \\
& \cellcolor{blue!10}\cmark
& \cellcolor{blue!10}\cmark
& \cellcolor{blue!10}\textbf{66.4}
& \cellcolor{blue!10}\textbf{48.4}
& \cellcolor{blue!10}3.2
& \cellcolor{blue!10}9.6 \\
\bottomrule
\end{tabular*}
\end{table}

% ----------------------- Table2 -------------------- %
\begin{table}[t]
\centering
\caption{Component and deployment ablations on ATR-UMOD.}
\label{tab:deploy_ablation_deployment}
\small
\setlength{\tabcolsep}{2pt}
\renewcommand{\arraystretch}{0.8}
\resizebox{\columnwidth}{!}{%
\begin{tabular}{@{}lcccccc@{}}
\toprule
Deployment & LCH & DCCM
& mAP$_{50}$
& mAP
& Params
& FLOPs\\
\midrule

\multirow{4}{*}{P3/P4/P5}
& \xmark & \xmark & 62.2 & 45.3 & 3.9 & 9.6 \\
& \cmark & \xmark & 63.7 & 46.7 & 4.3 & 10.5 \\
& \xmark & \cmark & 64.2 & 46.5 & 4.0 & 9.2 \\
& \cmark & \cmark & 65.4 & 48.0 & 4.7 & 11.1 \\
\midrule

\makecell[l]{P3/P4 (CrossP5)}
& \cellcolor{blue!12}\cmark
& \cellcolor{blue!12}\cmark
& \cellcolor{blue!12}\textbf{66.4}
& \cellcolor{blue!12}\textbf{48.4}
& \cellcolor{blue!12}3.2 ($-1.5$)
& \cellcolor{blue!12}9.6 ($-1.5$) \\
\bottomrule
\end{tabular}}
\end{table}
% ----------------------- Table3 -------------------- %
% \begin{table}[t]
% \centering
% \caption{\textbf{Internal ablation study of DCCM on DroneVehicle.}
% CMI and RC denote cross-modal injection and residual correction,
% respectively. Adaptive $\gamma$ indicates discrepancy-conditioned
% prediction of the correction strength; otherwise, $\gamma=\mathbf{1}$
% when RC is enabled.}
% \label{tab:dccm_ablation}
% \small
% \setlength{\tabcolsep}{3pt}
% \renewcommand{\arraystretch}{0.8}
% \begin{tabular*}{\columnwidth}{@{\extracolsep{\fill}}ccccccc@{}}
% \toprule
% CMI & RC & Adaptive $\gamma$
% & mAP$_{50}$ & mAP$_{75}$ & mAP & $\Delta$mAP \\
% \midrule
% \cmark & \cmark & \cmark
% & \textbf{82.4} & \textbf{78.6} & \textbf{67.0} & \textbf{+1.6} \\
% \cmark & \cmark & \xmark
% & 82.1 & 78.2 & 66.8 & +1.4 \\
% \cmark & \xmark & \xmark
% & 81.8 & 77.6 & 66.0 & +0.6 \\
% \xmark & \cmark & \xmark
% & 80.9 & 76.9 & 65.6 & +0.2 \\
% \xmark & \xmark & \xmark
% & 80.9 & 76.8 & 65.4 & +0.0 \\
% \bottomrule
% \end{tabular*}
% \end{table}
% ----------------------- Table4 -------------------- %
\begin{table}[t]
\centering
\caption{LCH mechanism analysis on DroneVehicle. All variants use the P3/P4 (CrossP5) deployment without DCCM.}
\label{tab:lch_mechanisms}
\small
\setlength{\tabcolsep}{3pt}
\renewcommand{\arraystretch}{0.8}
\begin{tabular*}{\columnwidth}{@{\extracolsep{\fill}}lcccc@{}}
\toprule
Variant
& mAP$_{50}$
& mAP
& Params
& FLOPs \\
\midrule
Local cross-attention & 81.5 & 65.9 & 3.0 & 8.4 \\
\midrule
w/o prototype guidance & 81.8 & 66.3 & 3.0 & 9.2 \\
w/o candidate return   & 81.8 & 66.4 & 3.1 & 9.3 \\
\textbf{LCH} & \textbf{82.0} & \textbf{66.6} & 3.1 & 9.3 \\
\bottomrule
\end{tabular*}
\end{table}

\begin{table}[t]
\centering
\caption{Sensitivity to prototype-guidance configuration on DroneVehicle. $M$ denotes the number of prototypes, and $\eta_c^{(0)}$ denotes the initial prototype-guidance strength.}
\label{tab:lch_configuration}
\small
\setlength{\tabcolsep}{3pt}
\renewcommand{\arraystretch}{0.8}
\begin{tabular*}{\columnwidth}{@{\extracolsep{\fill}}cccccc@{}}
\toprule
$M$ & $\eta_c^{(0)}$ & mAP$_{50}$ & mAP & Params & FLOPs \\
\midrule
2 & 0.25 & 82.7 & 67.2 & 3.17 & 9.6 \\
4 & 0.25 & \textbf{82.9} & \textbf{67.4} & 3.18 & 9.6 \\
6 & 0.25 & 82.1 & 66.9 & 3.19 & 9.6 \\
\midrule
4 & 0.50 & 82.6 & 67.2 & 3.18 & 9.6 \\
\bottomrule
\end{tabular*}
\end{table}
%%%%%%%%%%%%%%%%%%%%%%%%%%%%%%%%%%%%%%%%%%%%%%%%%%%%%%%%%%%%%%%%%%%%%%%%%%%%%%%
\noindent\textbf{Mechanism analysis of LCH.}
Table~\ref{tab:lch_mechanisms} investigates the contributions of prototype guidance and candidate return under the P3/P4 (CrossP5) deployment with DCCM disabled. LCH achieves 82.0\% mAP$_{50}$ and 66.6\% mAP, outperforming local cross-attention by 0.5 and 0.7 percentage points, respectively. Removing prototype guidance leads to drops of 0.2 and 0.3 points, confirming its benefit for local correspondence modeling. Disabling candidate return also reduces the two metrics by 0.2 points, highlighting the importance of propagating hyperedge messages back to candidate nodes for many-to-many cross-modal interaction.

Table~\ref{tab:lch_configuration} further examines the effects of prototype number and guidance strength. Increasing the prototype number from $M=2$ to $M=4$ improves mAP from 67.2\% to 67.4\%, while further increasing it to $M=6$ reduces mAP to 66.9\%. With $M=4$, increasing the initial prototype-guidance strength $\eta_c^{(0)}$ from 0.25 to 0.50 also reduces mAP to 67.2\%. We therefore adopt $M=4$ and $\eta_c^{(0)}=0.25$ as the default configuration.
% ----------------------- Table5 -------------------- %
%%%%%%%%%%%%%%%%%%%%%%%%%%%%%%%%%%%%%%%%%%

\noindent\textbf{Robustness to Cross-Modal Discrepancies.}
Following the public implementation of CBC-SWCA \citep{zhang2025mitigating}, we select 3,243 DroneVehicle test pairs with low RGB--T annotation similarity, measured using class-wise Hungarian-matched rotated-box IoU. This subset contains image pairs with greater spatial discrepancies and potential annotation inconsistencies between the two modalities. As shown in Table~\ref{tab:low_similarity}, our AP$_{50}$/AP gains over the baseline increase from 1.98/2.08 percentage points on the full test set to 2.72/2.82 points on the low-similarity subset.

\begin{table}[t]
\centering
\caption{Robustness evaluation on DroneVehicle. Full and Low-sim denote the full test set and low-similarity subset, respectively.}
\label{tab:low_similarity}
\footnotesize
\setlength{\tabcolsep}{3pt}
\renewcommand{\arraystretch}{0.8}
\begin{tabular*}{\columnwidth}{@{\extracolsep{\fill}}lcccc@{}}
\toprule
& \multicolumn{2}{c}{Full test set}
& \multicolumn{2}{c}{Low-sim subset} \\
\cmidrule(lr){2-3}\cmidrule(lr){4-5}
Method
& AP$_{50}$ & AP
& AP$_{50}$ & AP \\
\midrule
Baseline
& 80.9 & 65.3
& 76.7 & 61.9 \\
\textbf{Ours}
& \textbf{82.9} {\scriptsize(+2.0)}
& \textbf{67.4} {\scriptsize(+2.1)}
& \textbf{79.4} {\scriptsize(+2.7)}
& \textbf{64.8} {\scriptsize(+2.9)} \\
\bottomrule
\end{tabular*}
\end{table}

%%%%%%%%%%%%%%%%%%%%%%%%%%%%%%%%%%%%%%%%%%%%%%%%%%%%%%%%%%%%%%%
\section{Conclusion}
In this work, we presented HyperUAV, a hypergraph-guided framework for weakly aligned RGB--Thermal oriented object detection. To explicitly model uncertain local cross-modal correspondences, we introduced LCH, which constructs prototype-guided local hyperedges with adaptive soft associations, enabling bidirectional many-to-many information propagation between modalities. We further developed DCCM to perform reliability-aware cross-modal information injection and discrepancy-conditioned residual correction, selectively exploiting complementary information while preserving modality-specific characteristics. Extensive experiments on three UAV-based benchmarks, including DroneVehicle, ATR-UMOD, and ESCVehicle, demonstrate the effectiveness of HyperUAV, achieving superior detection performance with favorable accuracy-efficiency trade-offs.

% =========================================================
% LLM Usage Statement
% =========================================================
\clearpage
\section*{LLM Usage Statement}
During the preparation of this manuscript, ChatGPT (OpenAI) was used to assist with English language editing, including grammar correction and improving the clarity and readability of the manuscript. We did not use generative AI tools for tasks requiring mandatory disclosure, including research ideation, methodology development, experimental design, implementation of research methods, or data analysis. All AI-assisted content was carefully reviewed and
revised by the authors. We take full responsibility
for the final content of this work.
%-------------------------------------------------------------------------

%% file: sec/X_suppl.tex
\clearpage
\setcounter{page}{1}
\maketitlesupplementary

\section{Additional Experimental Results}
\label{sec:additional_results}

\subsection{DCCM Component Ablation}
Table~\ref{tab:dccm_internal_ablation} examines the contribution of DCCM on DroneVehicle with LCH disabled. Starting from 65.5\% mAP without DCCM, cross-modal injection alone reaches 65.7\%, while residual correction alone with $\gamma=1$ reaches 66.1\%. The full DCCM used in our model combines cross-modal injection with discrepancy-conditioned residual correction and achieves 66.3\% mAP. Relative to the baseline, it improves mAP$_{50}$, mAP$_{75}$, and mAP by 0.8 percentage points each.

\begin{table}[h]
\centering
\caption{\textbf{DCCM component ablation on DroneVehicle.}
All variants use the P3/P4 deployment with CrossP5 and LCH disabled.
CMI denotes cross-modal injection and RC denotes residual correction.}
\label{tab:dccm_internal_ablation}
\small
\setlength{\tabcolsep}{14pt}
\begin{tabular}{lccc}
\toprule
Variant & mAP$_{50}$ & mAP$_{75}$ & mAP \\
\midrule
w/o DCCM & 81.1 & 77.0 & 65.5 \\
+ CMI only     & 81.1 & 77.1 & 65.7 \\
+ RC only      & 81.5 & 77.5 & 66.1 \\
Full DCCM    & \textbf{81.9} & \textbf{77.8} & \textbf{66.3} \\
\bottomrule
\end{tabular}
\end{table}

\subsection{Robustness to Positional Shift}
\label{sec:rationale}
\begin{figure}[!t]
\centering
\includegraphics[width=\columnwidth]{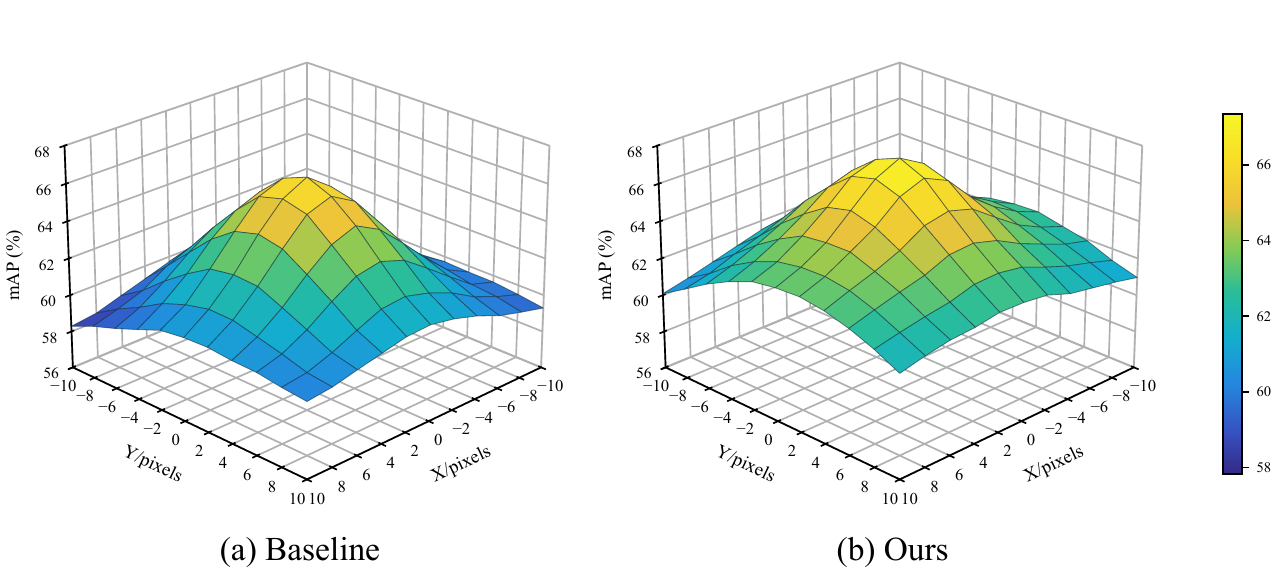}
\caption{Surface plot visualization of the position shift experiments. Horizontal coordinates indicate various shift steps along the x-axis and y-axis. Vertical coordinate denotes the detection performances in terms of mAP.}
\label{fig:position_shift}
\end{figure}

To evaluate the model’s robustness against weak misalignment quantitatively, we conduct experiments on DroneVehicle by manually simulating positional offsets~\citep{chen2024weakly,yang2026crossweaver}. First, we select 1,525 test pairs with identical RGB and IR box annotations, so that the annotated object locations coincide before translation. Then, keeping the IR image and its ground-truth boxes fixed, we shift the RGB image by $(\Delta x,\Delta y)$, with each component ranging from $-10$ to $10$ pixels in 2-pixel increments. The models are trained on the original training image pairs and evaluated on the shifted image pairs. As shown in Fig.~\ref{fig:position_shift}, our model maintains higher mAP than the baseline throughout the tested shift range. As the RGB displacement increases, its mAP declines more gradually on average, indicating greater robustness to cross-modal misalignment under these conditions.

%%%%%%%%%%%%%%%%%%%%%%%%%%%%%%%%%%%%%%%%%%%%%%%%%%%%%%%%%%%%%
\subsection{Results on Different Conditions}
\label{sec:RDC}
Table~\ref{tab:atr_umod_all_conditions} extends the weather analysis in the main paper to 21 subsets spanning six ATR-UMOD test conditions. HyperUAV-S achieves the highest reported mAP$_{50}$ on 19 of the 21 subsets. Its largest gain over PCDF~\citep{chen2025fusion} occurs under dim illumination (78.2\% vs.\ 67.7\%), while PCDF retains a slight lead under normal and near-night illumination (0.3 and 0.5 percentage points, respectively).

\begin{table}[t]
\centering
\caption{\textbf{Detection performance across ATR-UMOD test conditions.}
Results are reported as mAP$_{50}$ (\%). AL, AN, TI, WE, IL, and SC
denote altitude, viewing angle, time of day, weather, illumination, and
scenario, respectively~\citep{chen2025fusion}. HyperUAV-S averages over all 11 classes, assigning zero to classes without ground-truth instances in a subset. The highest value in each row is in bold.}
\label{tab:atr_umod_all_conditions}
\footnotesize
\setlength{\tabcolsep}{2pt}
\renewcommand{\arraystretch}{1.0}
\begin{tabular*}{\columnwidth}{@{\extracolsep{\fill}}clccccc@{}}
\toprule
\multicolumn{2}{c}{Condition} & CALNet & OAFA & YOLOrs & PCDF & HyperUAV-S \\
\midrule
\multicolumn{2}{l}{Full test set} & 53.8 & 57.9 & 60.0 & 63.3 & \textbf{68.8} \\
\midrule
\multirow{2}{*}{AL} & $[0,120]$ & 55.3 & 59.4 & 61.8 & 67.0 & \textbf{69.8} \\
 & $(120,300]$ & 41.4 & 47.3 & 48.3 & 50.3 & \textbf{57.0} \\
\midrule
\multirow{2}{*}{AN} & $[0,30]$ & 51.0 & 58.2 & 58.5 & 60.4 & \textbf{67.1} \\
 & $(30,75]$ & 43.1 & 51.7 & 55.0 & 57.2 & \textbf{62.4} \\
\midrule
\multirow{3}{*}{TI} & Morning & 43.9 & 53.6 & 55.6 & 58.9 & \textbf{63.8} \\
 & Afternoon & 53.8 & 60.0 & 62.7 & 66.3 & \textbf{71.8} \\
 & Night & 31.2 & 38.2 & 35.3 & 41.7 & \textbf{45.2} \\
\midrule
\multirow{5}{*}{WE} & Cloudy & 54.8 & 62.1 & 69.6 & 71.2 & \textbf{78.1} \\
 & Foggy & 28.8 & 30.9 & 33.2 & 32.6 & \textbf{38.8} \\
 & Snowy & 38.2 & 48.4 & 50.1 & 53.4 & \textbf{56.0} \\
 & Sunny & 48.7 & 54.6 & 57.1 & 60.1 & \textbf{64.7} \\
 & Night & 32.5 & 38.0 & 35.4 & 41.3 & \textbf{44.8} \\
\midrule
\multirow{4}{*}{IL} & Normal & 49.8 & 55.0 & 56.9 & \textbf{60.6} & 60.3 \\
 & Dim & 52.5 & 59.8 & 67.5 & 67.7 & \textbf{78.2} \\
 & Near Night & 41.5 & 44.8 & 47.5 & \textbf{48.1} & 47.6 \\
 & Night & 31.5 & 37.3 & 34.6 & 40.0 & \textbf{44.8} \\
\midrule
\multirow{5}{*}{SC} & Construction site & 41.0 & 42.3 & 47.2 & 49.0 & \textbf{56.1} \\
 & Factory & 17.6 & 23.0 & 22.4 & 23.0 & \textbf{24.0} \\
 & Neighborhood & 33.7 & 34.6 & 31.9 & 35.3 & \textbf{36.7} \\
 & Parking Lot & 31.9 & 39.5 & 38.0 & 40.8 & \textbf{41.3} \\
 & Road & 49.6 & 57.0 & 59.3 & 62.3 & \textbf{69.1} \\
\bottomrule
\end{tabular*}
\end{table}

\begin{figure*}[!t]
\centering
\begin{minipage}[t]{0.48\textwidth}
    \centering
    \includegraphics[width=0.8\linewidth]{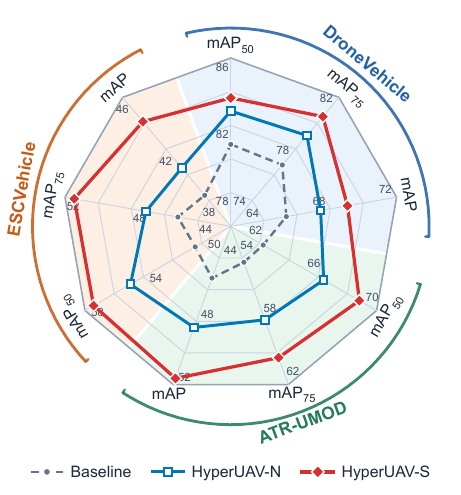}
    \caption{Comparison of detection performance among the baseline, 
    HyperUAV-N, and HyperUAV-S across three benchmark datasets.}
    \label{fig:radar}
\end{minipage}
\hfill
\begin{minipage}[t]{0.48\textwidth}
    \centering
    \includegraphics[width=\linewidth]{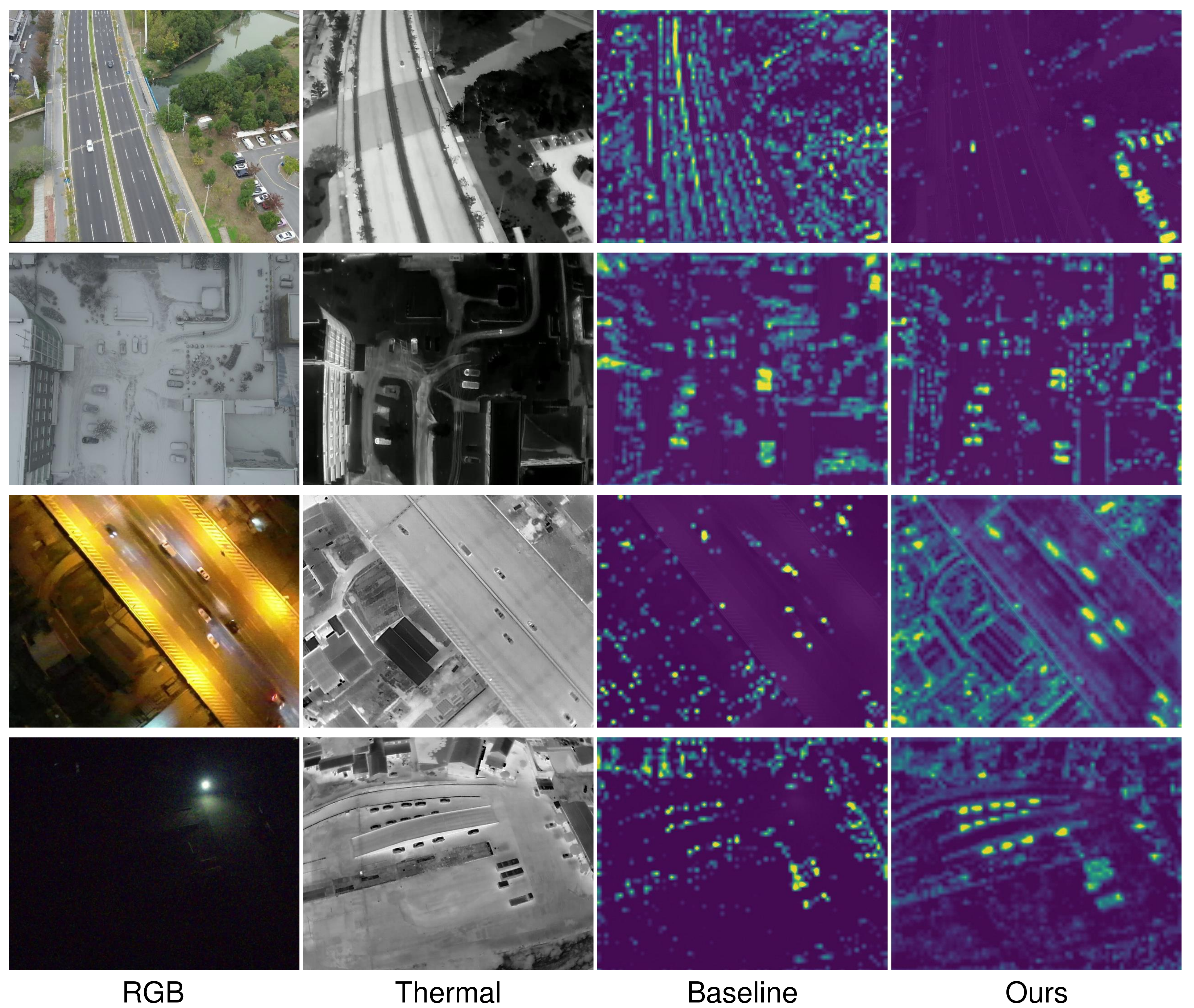}
    \caption{Qualitative comparison of fused feature responses 
    across diverse scenes. From left to right: RGB images, thermal 
    images, and feature heatmaps of the baseline and the proposed method.}
    \label{fig:heatmap}
\end{minipage}
\end{figure*}

\section{Visualization}
\label{sec:Vis}

Fig.~\ref{fig:radar} provides an overall comparison of detection 
performance on DroneVehicle, ESCVehicle, and ATR-UMOD. Both 
HyperUAV-N and HyperUAV-S consistently outperform the baseline 
across all three datasets in terms of mAP, mAP$_{50}$, and 
mAP$_{75}$, demonstrating the effectiveness and generalization 
capability of the proposed framework.

Fig.~\ref{fig:heatmap} further visualizes the fused feature 
responses under diverse imaging conditions, including daytime, 
nighttime, and low-light scenes. Compared with the baseline, 
HyperUAV generally exhibits stronger and more concentrated 
responses in target regions while reducing irrelevant background 
activations. These visualizations demonstrate its ability to 
exploit complementary cross-modal information and enhance 
target-related feature representations under challenging 
imaging conditions.

% \begin{figure}[t]
% \centering
% \includegraphics[width=.9\columnwidth]{Figures/fig-radar.pdf}
% \caption{Comparison of detection performance among the baseline, HyperUAV-N, and HyperUAV-S across three benchmark datasets.}
% \label{fig:radar}
% \end{figure}

% \begin{figure}[t]
% \centering
% \includegraphics[width=.9\columnwidth]{Figures/fig-heatmap.pdf}
% \caption{Qualitative comparison of fused feature responses across diverse scenes. From left to right: RGB images, thermal images, and feature heatmaps of the baseline and the proposed method.}
% \label{fig:heatmap}
% \end{figure}